\documentclass[conference]{IEEEtran}
\usepackage{cite}

\ifCLASSINFOpdf
\else
\fi
\usepackage{array}

\usepackage{stfloats}
\usepackage{url} 

\usepackage[usenames,dvipsnames]{pstricks}
\usepackage{epsfig}
\usepackage{pst-grad} 
\usepackage{pst-plot} 
\usepackage{float}

\usepackage{etoolbox}

\usepackage{hyperref}

\usepackage{multirow}

\begin{document}
%
\title{Iterative temporal differencing with random synaptic feedback weights support error backpropagation for deep learning}


\author{
    \IEEEauthorblockN{
    Aras R. Dargazany}
    \IEEEauthorblockA{
	Department of Electrical, Computer, and Biomedical Engineering, 
    University of Rhode Island, 
    Kingston, RI 02881, USA}
	\IEEEauthorblockA{Corresponding Author, arasdar@uri.edu}
}
\maketitle

\begin{abstract}
This work shows that a differentiable activation function is not necessary any more for error backpropagation. 
The derivative of the activation function can be replaced by an iterative temporal differencing using fixed random feedback alignment.
Using fixed random synaptic feedback alignment with an iterative temporal differencing
is transforming the traditional error backpropagation into a more biologically plausible approach for learning deep neural network architectures.
This can be a big step toward the integration of STDP-based error backpropagation in deep learning.
\end{abstract}


%
\IEEEpeerreviewmaketitle


\begingroup
\let\clearpage\relax

\section*{Introduction}

Vanilla backpropagation was proposed in 1987 \cite{rumelhart1986learning,rumelhart1985learning,rumelhart1986learning_2}.
Almost at the same time, biologically-inspired convolutional networks was introduced as well \cite{lecun1989backpropagation}.

Deep learning was introduced as an approach to learn deep neural network architecture using vanilla backpropagation \cite{lecun1989backpropagation,lecun2015deep,krizhevsky2012imagenet}.
Extremely deep networks learning reached 152 layers of representation with residual and highway networks \cite{he2016deep,srivastava2015highway}.

Deep reinforcement learning was successfully implemented and applied which was mimicking the dopamine effect in our brain for self-supervised and unsupervised learning \cite{silver2016mastering,mnih2015human,mnih2013playing}.

Hierarchical convolutional neural network have been biologically inspired \cite{hubel1959receptive,fukushima1988neocognitron,fukushima1975cognitron,yamins2016using}.

Geoff Hinton in 1988 proposed the temporal derivative in backprop (backpropagation) with recirculation \cite{hinton1988learning} which does not require the derivative of the activation function.
He gave a lecture about this approach again in NIPS 2007 \cite{hinton2007backpropagation}, and recently gave a similar lecture in Stanford in 2014 and 2017 to reject the four arguments against the biological foundation of backprop.

The discovery of fixed random synaptic feedback in error backpropagation in deep learning started a new quest of finding the biological version of backprop \cite{lillicrap2016random} since it solves the symmetrical weights or synapses problem in backprop.

Recently, spike-time dependent plasticity was the important issue with backprop.
Apical dendrite as the segregated synaptic feedback are claimed to be modeled into the backprop successfully \cite{guergiuev2016biologically}.

\begin{table}
\centering
\begin{tabular}{|p{0.3\textwidth}|p{0.2\textwidth}|}
\hline\hline 
\multicolumn{1}{|c|}{\multirow{1}{*}{\textbf{Problems}}} & \multicolumn{1}{|c|}{\textbf{Solutions}} \\
\cline{1-2} 
Hierarchical layers of representation & Vanilla backprop\cite{rumelhart1986learning} \\
\hline
Deep convolutional layers & Convolutional neural nets \cite{lecun1989backpropagation,lecun2015deep,yamins2016using,fukushima1988neocognitron,hubel1959receptive} \\
\hline
Extremely Deep networks & Residual and highway networks \cite{he2016deep,srivastava2015highway} \\
\hline
Dopamine & Deep reinforcement learning \cite{mnih2015human,silver2016mastering,mnih2013playing} \\
\hline
Spike & Dropout \cite{srivastava2014dropout} and linear-nonlinear-Poisson (LNP) models using Poisson distribution \\
\hline
spike time-dependent plasticity (STDP) (\textbf{Our core contribution}) & Segregated apical dendrites \cite{guergiuev2016biologically} \\
\hline
Symmetry or symmetrical neurons & Fixed random synaptic feedback alignment \cite{lillicrap2016random} \\
\hline
\hline
\end{tabular}
\caption{The problems with with artificial neural networks compared to the biological neural networks (brain) according to neuroscientists.}
\label{tab:intro}
\end{table}

\section*{Iterative temporal differencing} 

Iterative temporal differencing vs vanilla back prop and fixed random synaptic feedback alignment are illustrated in figure \ref{fig:prop}.

\begin{figure*}
\centering
\includegraphics[width=\linewidth]{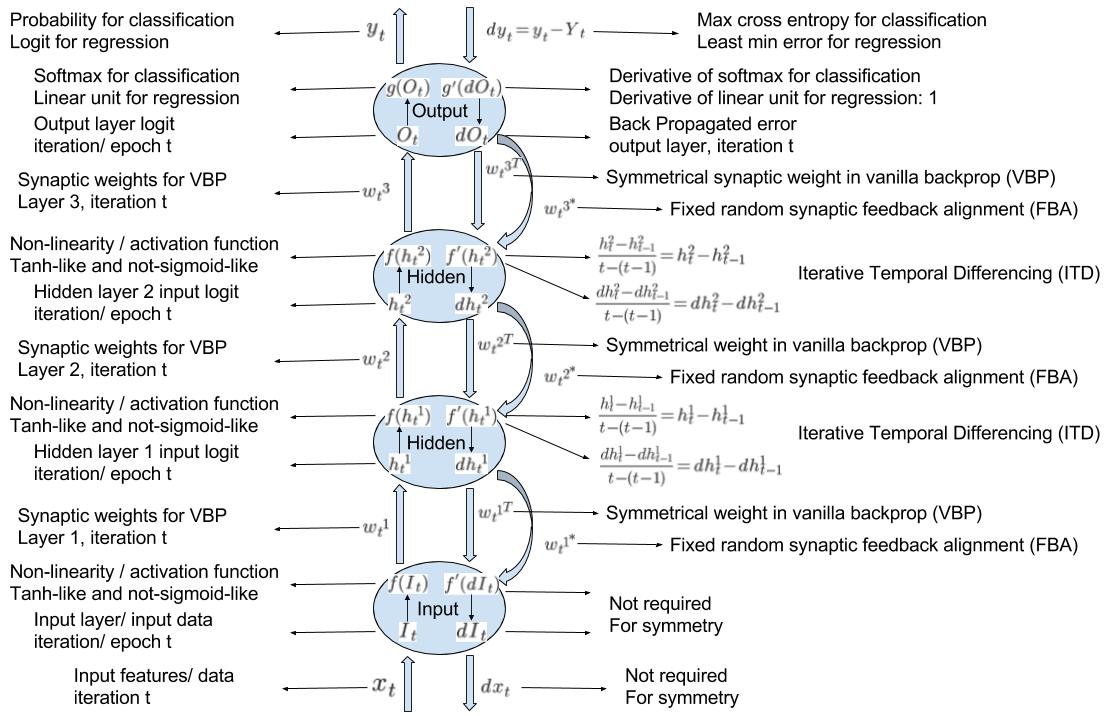}
\caption{Vanilla backprop vs feedback alignment vs iterative temporal differencing.}
\label{fig:prop}
\end{figure*}
\section*{Experimental results}

The experimental results of learning the same network are illustrated in this figure \ref{fig:exp} with these hyper-parameters:
\begin{itemize}
\item number of iterations or epochs: 100000
\item learning rate: 1e-3
\item minibatch size: 50
\item number of hidden units: 32
\item number of input and output units: MNIST standard dataset as (\url{http://yann.lecun.com/exdb/mnist/})
\item number of hidden layers
\end{itemize}



\begin{figure} 
\centering
\includegraphics[width=\linewidth]{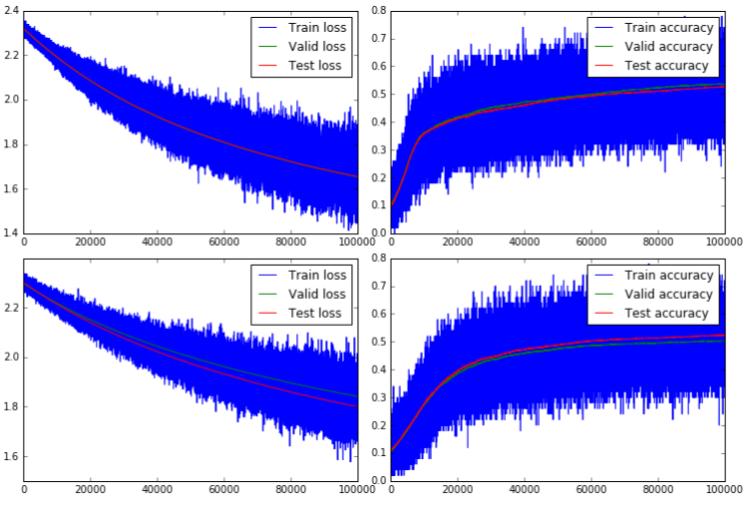}
\includegraphics[width=\linewidth]{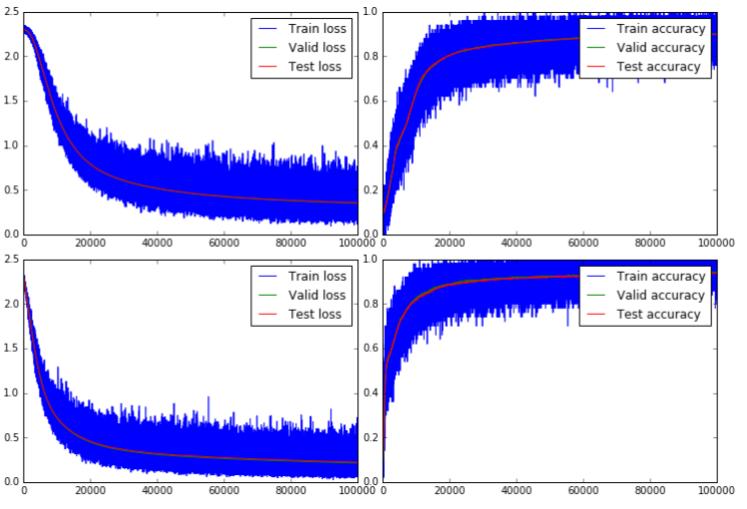}
\caption{The experimental results on MNIST dataset from top to bottom order:  FBA + ITD-y,  FBA + ITD-dy, FBA and VBP. Some acronyms: iterative temporal differencing (ITD}, feedback alignemtn (FBA), vanilla backprop (VBP) 
\label{fig:exp}
\end{figure}


\section*{Discussion \& future view}
In this paper, we took one more step toward a more biologically plausible backpropagation for deep learning.
After hierarchical convolutional neural network and fixed random synaptic feedback alignment, we believe iterative temporal differencing is a way toward integrating STDP learning process in the brain.
We believe the next steps should be to investigate more into the STDP processes details in learning, dopamine-based unsupervised learning, and generating Poisson-based spikes.
These steps are shown in this table \ref{tab:disc}.

\begin{table} 
\centering
\begin{tabular}{|p{0.4\columnwidth}|p{0.4\columnwidth}|}
\hline\hline 
\multicolumn{1}{|c|}{\multirow{1}{*}{\textbf{Problems}}} & \multicolumn{1}{|c|}{\textbf{Solutions}} \\
\cline{1-2} 
Very Deep Hierarchical convolutional layers of representation & solved \\
\hline
Dopamine & almost solved \\
\hline
Spike & Approached but not solved yet. Over fitting problem and extremely low-power consumption of the brain compared to deep learning machines. \\
\hline
STDP (\textbf{Our core contribution}) & Almost solved with our approach but should be experimented more and discussed more with neuroscientists. \\
\hline
Symmetry or symmetrical neurons & solved \\
\hline
STDP (\textbf{Our core contribution}) & Almost solved with our approach but should be experimented more and discussed more with neuroscientists. \\
\hline
A combined model of all these parameters & important future direction \\
\hline
\hline
\end{tabular}
\caption{The solved and unsolved problems with artificial neural networks are shown in order to give a clear future direction for understanding the mystery behind the learning process in our biological neural network.}
\label{tab:disc}
\end{table}
\endgroup


\end{document}